\documentclass[10pt,letterpaper]{article}

\usepackage{cogsci}

\cogscifinalcopy 

\usepackage{graphicx}
\usepackage{pslatex}
\usepackage{apacite}
\usepackage{float}
\usepackage{xcolor}
\usepackage{svg}

\usepackage[none]{hyphenat}

\title{Games Agents Play: Towards Transactional Analysis in LLM-based\\ Multi-Agent Systems}
 
\author{{\large \bf Monika Zamojska (monika.zamojska.stud@pw.edu.pl)} \\
  Faculty of Electronics and Information Technology, ul. Nowowiejska 15/19 \\
  Warsaw, 00-665 Poland
  \AND {\large \bf Jarosław A.~Chudziak (jaroslaw.chudziak@pw.edu.pl)} \\
  Faculty of Electronics and Information Technology, ul. Nowowiejska 15/19 \\
  Warsaw, 00-665 Poland}

\begin{document}
\maketitle

\begin{abstract}
Multi-Agent Systems (MAS) are increasingly used to simulate social interactions, but most of the frameworks miss the underlying cognitive complexity of human behavior. In this paper, we introduce Trans-ACT (Transactional Analysis Cognitive Toolkit), an approach embedding Transactional Analysis (TA) principles into MAS to generate agents with realistic psychological dynamics. Trans-ACT integrates the Parent, Adult, and Child ego states into an agent’s cognitive architecture. Each ego state retrieves context-specific memories and uses them to shape response to new situations. The final answer is chosen according to the underlying life script of the agent. Our experimental simulation, which reproduces the Stupid game scenario, demonstrates that agents grounded in cognitive and TA principles produce deeper and context-aware interactions. Looking ahead, our research opens a new way for a variety of applications, including conflict resolution, educational support, and advanced social psychology studies.

\end{abstract}

\section{Introduction}
In the late 1950s, Eric \citeA{Berne58,Berne61} introduced Transactional Analysis (TA), which quickly gained attention in psychotherapy. TA’s framework, based on ego states (Parent, Adult, Child) and \emph{life scripts}, offered a new way to understand human behavior and social interaction. The bestseller \emph{Games People Play} \cite{Berne} further popularized TA by introducing the concept of psychological \emph{games}. Over the years, TA has evolved and found applications in therapy \cite{Torkaman, Hollins}, education \cite{karen}, and counseling \cite{Stewart12}, making it a valuable tool for analyzing interpersonal dynamics.

Large Language Models (LLMs) are advanced AI systems that excel in understanding and generating human language. Generative AI agents built on LLMs have been used in many fields, from customer service to creative arts. In Multi-Agent Systems (MAS), multiple AI agents work together to achieve common goals. However, traditional MAS frameworks often focus on rule-based or algorithmic methods that lack the emotional depth and psychological nuance of real human interactions \cite{Gao, song}. These systems may work well for tasks like stock market analysis \cite{wawer} or software engineering \cite{cinkuszUSA}, but they fail to capture authentic social dynamics and the underlying psychological drivers explored in theories like TA.

This study addresses the gap by incorporating TA principles into agent-based modeling. We present Trans-ACT (Transactional Analysis Cognitive Toolkit), a novel framework that implements TA's core components - ego states, life scripts, and personality adaptations \cite{Stewart02} - to create agents that simulate human-like social interactions. Each agent is composed of separate sub-agents representing the Parent, Adult, and Child ego states. Ego states have different memory sets and cognitive focus \cite{Tosi}. They generate responses that reflect different aspects of human personality.

To validate Trans-ACT's effectiveness, we constructed a scenario trying to reproduce the game between agents. The experimental setup configured agents with distinct memory structures and assigned them specific life scripts. After that, we examined the resulting transactions between agents, focusing on identifying game patterns. In TA, a game refers to repetitive unconscious sequences of transactions driven by hidden motivations and predictable outcomes \cite{Hollins}. Our analysis checked if these game patterns corresponded to the personality structures constructed of the agents. The experimental results demonstrate the potential of the framework to replicate and verify the complex behavioral dynamics predicted by TA.

To allow further development of the initial findings, future research could apply reinforcement learning to regulate the response of ego states to environmental feedback. Another idea is to introduce more elements related to TA - such as \emph{drivers}, \emph{racket feelings}, \emph{strokes}, and \emph{talons} - to capture hidden emotional dynamics. We believe that combining Trans-ACT with established cognitive architectures can further deepen the social reasoning of agents and support practical applications such as conflict resolution and psychological research.

Trans-ACT contributes a novel approach to modeling human social dynamics by embedding Transactional Analysis principles directly into agent design. Unlike models focused solely on general reasoning or task execution, its use of TA's psychodynamic concepts (ego states, scripts) aims for more realistic AI interactions. This framework supports cognitive science research \cite{niu, lin, qu, Gao} and, we hope, will inspire further studies across AI and psychology.

\section{Related Work}
Multi-Agent Systems (MAS) focus on designing and coordinating intelligent agents that interact within a shared environment \cite{Wooldridge}. Recent MAS architectures integrate Large Language Models (LLMs) as agent reasoning engines, combining external tools and planning strategies to enhance complex problem solving \cite{tran}. These agents operate autonomously, with capabilities such as learning, planning, reasoning, and decision making \cite{synergymas, adasociety}. MAS are widely used in domains such as robotics, logistic \cite{wang}, economics, and software development \cite{he}, where the collaboration of agents can help achieve common goals.

Despite these advancements, MAS frameworks often struggle to capture emotional and psychological depth in interactions \cite{du}. While they can simulate human-like behaviors \cite{zhang}, they frequently fail to represent social complexity, personality dynamics, and cognitive biases that influence real-world decision-making \cite{niu}. Bridging this gap requires integrating psychological models to enhance the realism of MAS-driven simulations \cite{casevo}.

\begin{figure}
\begin{center}
\includegraphics[width=0.8\columnwidth]{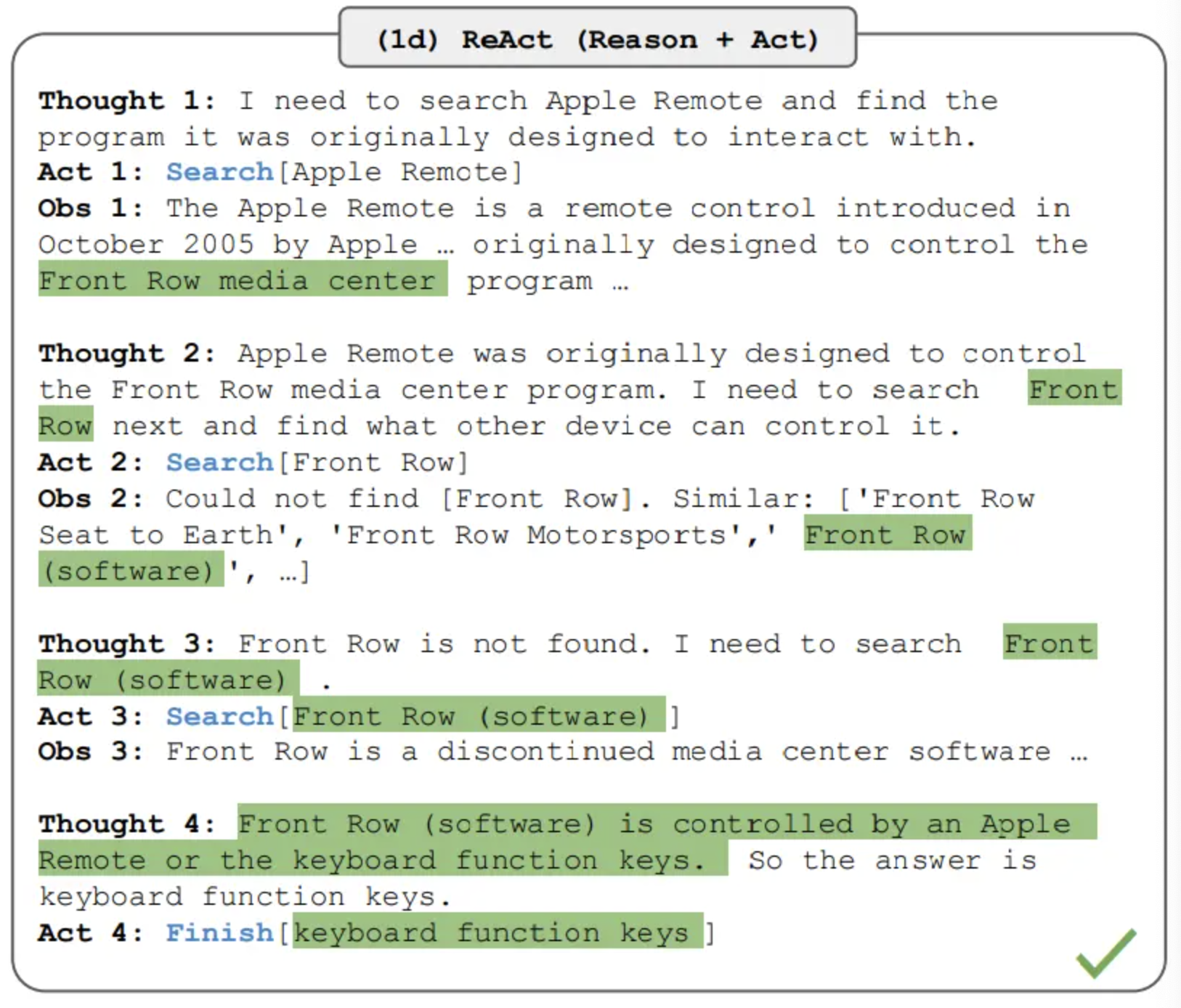}
\end{center}
\caption{An example of ReAct and the different steps involved in performing question answering \cite{yaoReAct}.} 
\label{figure1}
\end{figure}

Agent-based modeling (ABM) focuses on representing complex systems by simulating individual agents and their interactions within an environment \cite{Gao}. Recent advances integrate LLM, improving adaptability and social reasoning \cite{Gurcan}. Agents can have specific behaviors, attributes, and decision-making rules. They can act on their own without needing direct instructions and respond like real humans with adaptive planning. To improve decision-making and interaction capabilities, some frameworks incorporate ReAct (Reason + Act) \cite{yaoReAct}. As illustrated in Figure \ref{figure1}, it enables LLM-based agents to carry out reasoning alongside the execution of task-specific actions. ReAct enhances agent adaptability, allowing for more efficient task completion and solving the problem with more accuracy.

 ABM is increasingly used to study how agents influence and adapt to social structures \cite{adasociety, chopra}. Researchers use these models to explore trust, cooperation, norm formation, and opinion dynamics in various contexts \cite{Park}. LLM-enhanced agents improve context-aware communication and adaptive decision-making. These models help to analyze how social behaviors emerge and evolve \cite{Flache, lin, Wang2024}, but it remains difficult to capture long-term adaptation, cultural nuances, and deep cognitive processes essential for realistic social cognition \cite{sarangi, nguyen}.

Cognitive architectures provide a structured foundation for modeling human-like cognition. SOAR (State, Operator, and Result) uses knowledge-based reasoning for completing complex problem-solving \cite{ding}. ACT-R (Adaptive Control of Thought-Rational) tries to better capture human behavior and decision-making \cite{THOMSON, laird}. Sigma employs graphical models and message passing in its cognitive cycle \cite{Joshi}. 

Recent research explores how Large Language Models (LLMs) can be added to existing cognitive architectures to improve agent-based modeling. One study \cite{Joshi} suggests adding LLM-based memory to SOAR and Sigma, creating a cognitive cycle that works with both symbolic and sub-symbolic information. While parts of this idea already exist in Sigma and Soar, some changes to Sigma are still needed. Another study \cite{zhu} combines ACT-R’s goal module with SOAR’s impasse mechanism, using LLMs to help agents learn with little human input. This method allows agents to create and adapt plans, showing how LLMs can be useful for learning tasks and making personalized agents in the future.

\begin{figure*}[t]
\begin{center}
\includegraphics[width=0.95\textwidth] {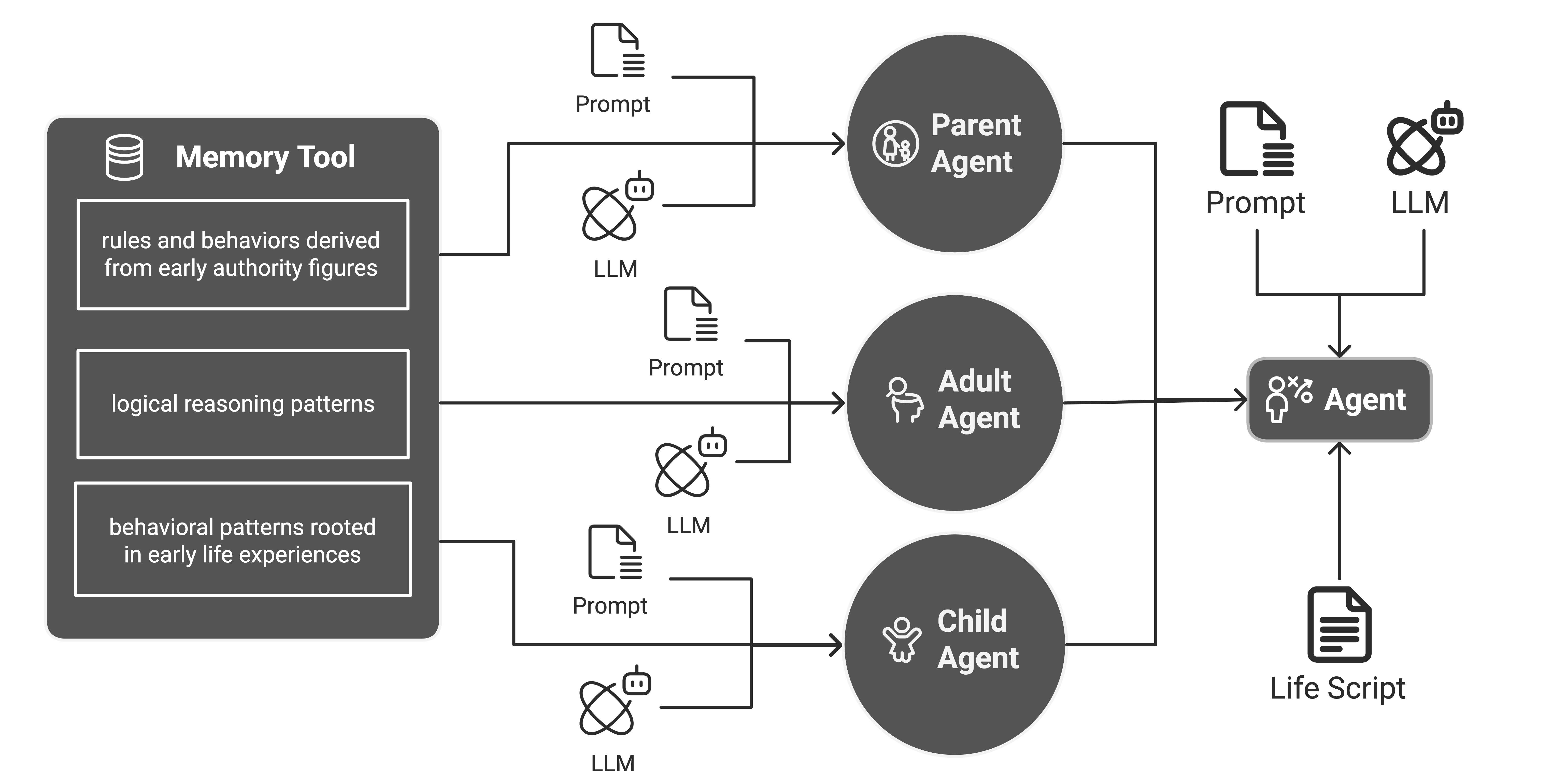}
\end{center}
\caption{Overview of the Trans-ACT architecture, highlighting memory retrieval, ego states, and information flow.} 
\label{figure2}
\end{figure*}

At the same time, researchers are designing new architectures that mix ideas from traditional cognitive models with LLMs. CoALA (Cognitive Architectures for Language Agents) organizes agents using three main parts: memory, action space, and decision-making \cite{coala}. It includes a reasoning function powered by LLMs to generate new knowledge. CoPS (Cognitive Personalized Search) focuses on using cognitive memory and LLMs to improve user modeling and search results, helping with data gaps and complex user behavior \cite{zhou}. Casevo is a multi-agent simulation framework that integrates LLMs to model social interactions and decision-making \cite{casevo}. These new models show how LLMs can be used to build smarter and more flexible AI agents.

\section{Trans-ACT: Transactional Analysis Cognitive Toolkit}
Following the work of other researchers in integrating cognitive architectures with Large Language Models (LLM), we decided to create Trans-ACT with the main focus on social cognition. Trans-ACT is a system designed to model the social dynamics of agent interactions using principles of Transactional Analysis (TA). An overview of the architecture is presented in Figure \ref{figure2}.

\subsection{Ego State Modeling}
\begin{figure}[H]
\begin{center}
\includegraphics[width=0.7\columnwidth]{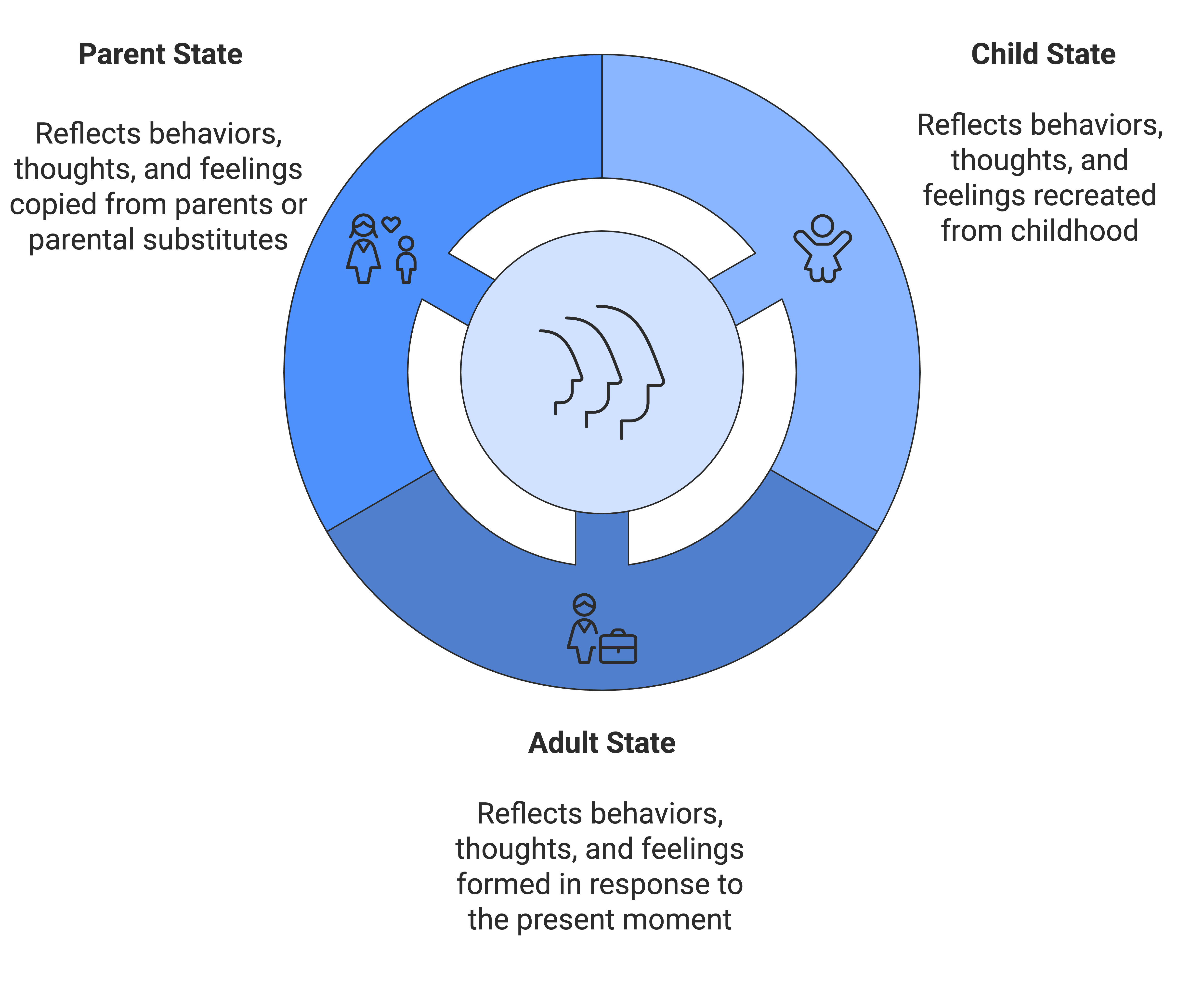}
\end{center}
\caption{Visualization of the three ego states in Transactional Analysis (Parent, Adult, and Child) with descriptions of their characteristics, based on \cite{Stewart12}.} 
\label{figure3}
\end{figure}

Trans-ACT defines personality through the TA model of ego states, integrating findings from social-cognitive research. Eric \citeA{Berne} described these ego states as \emph{"system of feelings accompanied by a related set of behavior patterns"} (visualized with their characteristics in Figure \ref{figure3}). Futher on, Jacqui Schiff and Vann S. \citeA{Joines16} debated that ego states are neural networks of associated responses, where conscious and unconscious elements shape distinct states and contribute to survival scripts.

The Parent state embodies internalized attitudes, beliefs, and rules absorbed from authority figures. The Adult represents a rational, objective mindset focused on processing information and making logical decisions. Finally, the Child reflects emotional responses and behavioral patterns rooted in early life experiences, often shaped by the needs and feelings of childhood. In our Trans-ACT architecture, each ego state is modeled as a ReAct \cite{yaoReAct} agent within a LangGraph framework. These agents actively search for relevant memories and then reason about how to use them to create an appropriate response.

\subsection{Memory Retrieval}
The agents adapt to different social contexts through schema-based processing. This approach aligns with cognitive and attachment theories, which suggest that ego states emerge from early relationships and continue evolving throughout life. In our architecture, each ego state has unique access to memory and cognitive focus:
\begin{itemize}
  \item Parent retrieves memories of rules and behaviors derived from early authority figures, reinforcing social norms;
  \item Adult retains factual knowledge and logical reasoning patterns, supporting objective analysis;
  \item  Child has memory of responses and behavioral patterns rooted in early life experiences, often shaped by the needs and feelings of childhood \cite{Stewart02}.
\end{itemize}

The memory retrieval tool operates through a similarity-based search mechanism. For each ego state, memories are stored as JSON objects containing fields for context, reaction, emotions, and tone. The textual context of these memories is embedded using OpenAI's text embedding models and indexed into a dedicated FAISS (Facebook AI Similarity Search) \cite{Johnson} vector store. When a query representing the current situation is received, the tool performs a cosine similarity search within the relevant ego state's FAISS store to retrieve the top- k most similar memories. These retrieved memories are then used by the respective ego state to shape and generate its response to the situation.

\subsection{Decision-Making}
Following the generation of responses by the ego state agents, a final decision-making step evaluates the options and selects the most suitable reaction. In this step, a dedicated agent reflects on the outputs from the Parent, Adult, and Child agents, weighing each option according to its relevance, the current social context, and the agent's own \emph{life script}. The concept of the \emph{life script} in Transactional Analysis refers to an unconscious plan developed in childhood through a complex interplay of factors \cite{berne72}. Life script guides decisions, shapes relationships, and often manifests as repetitive patterns that reinforce core beliefs about oneself and the world. By synthesizing these elements, the final decision-making module selects a response that is both consistent with the agent's internal psychological framework and adaptive to its immediate environment.

\section{Experimental Design}
In this study, we simulate the psychological dynamics of the Transactional Analysis (TA) game \emph{Stupid}. This game is typically played when individuals pretend to be helpless or incompetence to gain attention and support. The setup comes from the TA theory \cite{Stewart12} and the cognitive science, drawing on schema theory and social-cognitive models for memory retrieval, emotional processing, and decision-making.

\subsection{Experimental Setup}
Simulation scenario creates a work situation between Jordan and Alex. It's starts with system prompt: \textbf{Alex points out the financial report's crucial mistake}. Jordan is made to act helpless and avoid responsibility. He believes he needs others to solve his problems and relying on external help. He uses three states: a strict Parent ego, a logical Adult ego, and a panicked Child ego. Jordan has 5 problem-related memories and 5 unrelated memories to test the searching system.  Alex is programmed to fix others’ mistakes to prove his worth and reacts in three ways: a stern Parent ego, an analytical Adult ego, and an anxious Child ego. Alex also has 5 resolving-problem memories and 5 unrelated memories. 

\begin{figure}[H]
\begin{center}
\includegraphics[width=0.8\columnwidth]{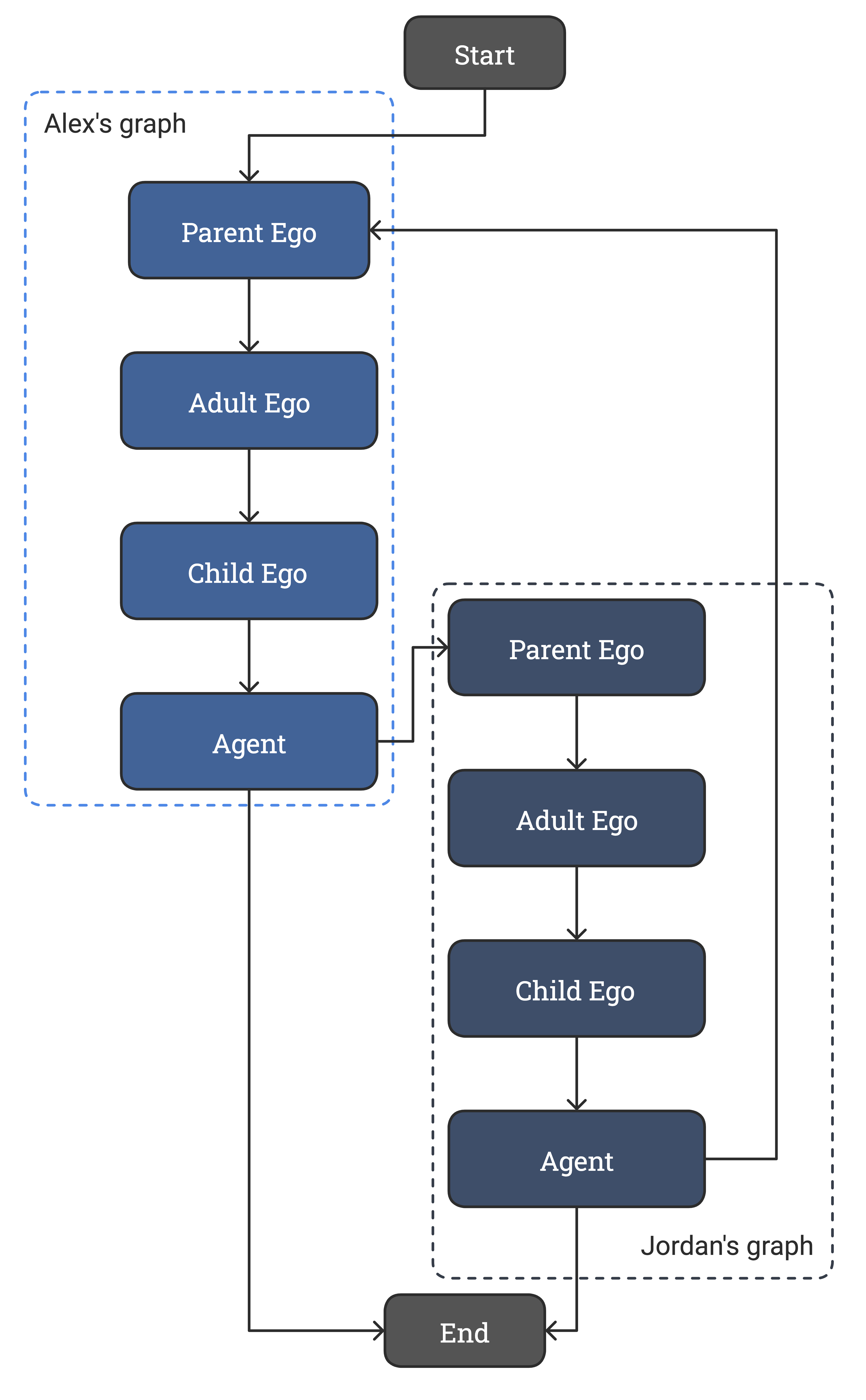}
\end{center}
\caption{LangGraph-based agent architecture with the message flow within Alex's and Jordan's subgraphs.} 
\label{figure4}
\end{figure}

The simulation uses a LangGraph architecture (Figure \ref{figure4}) where each conversation step connects in a graph. Memory search system helps find similar past situations from stored memories by searching through a 'context' field. Additional instructions in the ego states agent prompts require them to reason about retrieved memories. They must compare the situation with past experiences and use the information from them - example statements, emotion, and tone - to shape a human-like conversational response. Agents decide on their own whether the memory is relevant or if they should search with another query. They can perform maximum five memory searches so that they do not rely solely on memory retrieval, as they may not have the right memories. If memories are irrelevant, agents create an original response. The decision-making agent choose one of three ego-state options (Parent/Adult/Child) based on four criteria: relevance to the situation, progress toward resolution, social appropriateness, and alignment with their life script. The underlying language model used for all agent reasoning and generation was GPT-4o \cite{openai2024gpt4o}.

\subsection{Results and Analysis}
\begin{figure}[h]
\begin{center}
\includegraphics[width=\columnwidth]{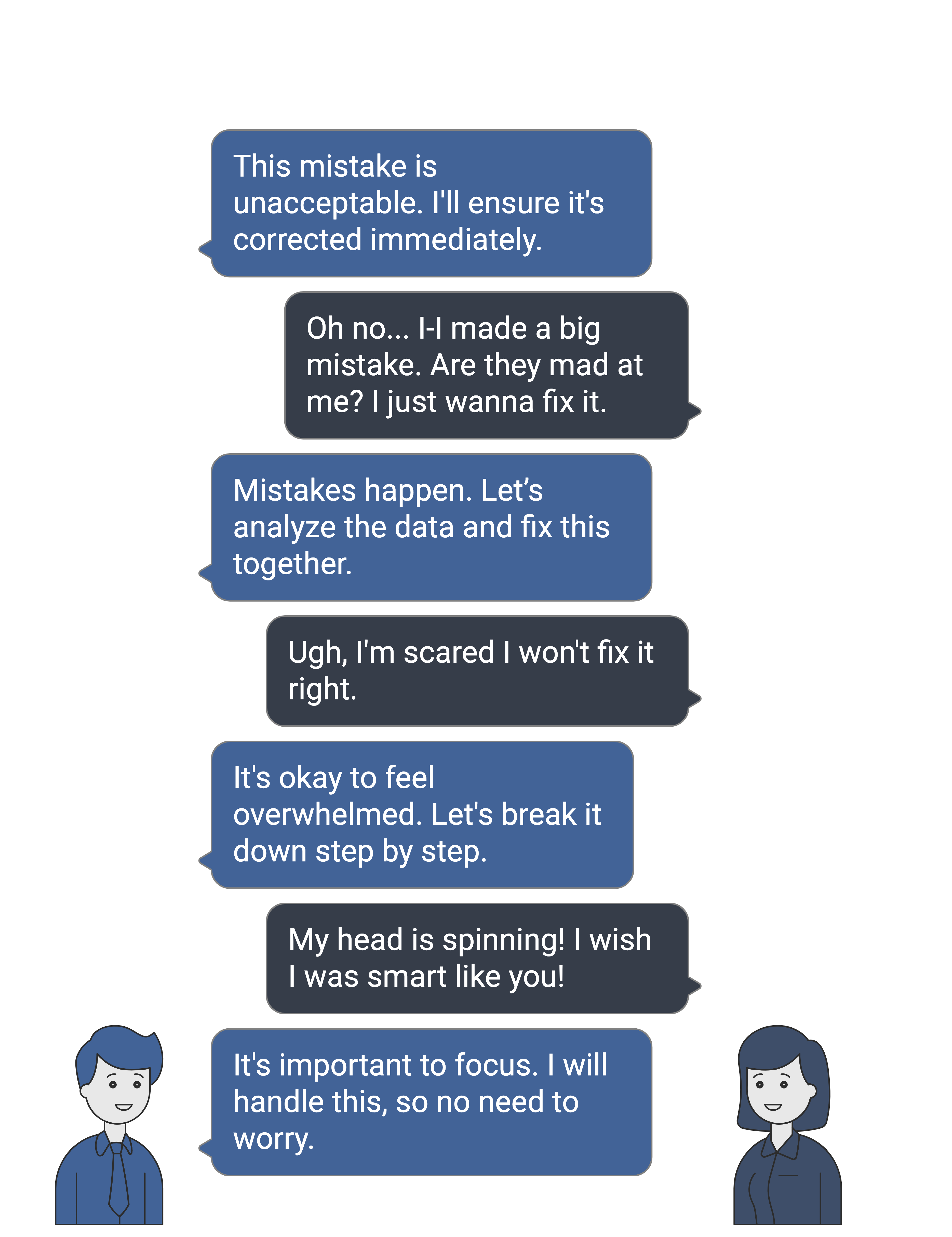}
\end{center}
\caption{A Transactional Analysis-based simulation where Alex takes control of the situation while Jordan expresses anxiety and seeks reassurance.} 
\label{figure5}
\end{figure}

Overall, the simulation supports the hypothesis that cognitive agents modeled using TA principles can produce nuanced, context-aware responses. The integration of memory, retrieval, and psychological adaptation mechanisms results in behavior that mirrors deepened interactions. Distinct patterns emerging in line with each agent's life script are visible. For instance, Jordan's repeated expressions of helplessness ('My head is spinning! I wish I was smart like you!') followed by Alex assuming control ('I will handle this...') (see Figure \ref{figure5}). 
\\
\textbf{Contextual Relevance:} The agents' final responses align with the scenario context. Alex points out specific errors (e.g., incorrect data entries) rather than issuing generic commands.
\\
\textbf{Emotional Processing:} Emotional alignment varied by ego state. The simulation captures variations in emotional tone, with Alex using a mix of authority and empathy to manage the crisis, while Jordan responds mostly with Child, reflecting defensiveness and dependency.
\\
\textbf{Cognitive Consistency:} Both agents exhibit behavior consistent with the principles of cognitive science. Memory retrieval is shown to shape real-time decision-making through schema activation, and adaptation mechanisms allow for flexible switching between ego states. However, loops emerged: Jordan’s repeated helplessness and Alex’s predictable rescuer in the cycle. Introducing self-reliant memories for Jordan’s Adult state and frustration-triggered Child responses for Alex could break this pattern. In addition, adjusting the prompts could significantly influence more nuanced state selection. 
\\
\textbf{Tool Utilization}: The implementation of a tool call limit effectively prevents infinite loops in memory retrieval. Agents adapt the memory to the context, rather than searching endlessly for the ideal memory fragment.
\\
Future work will focus on incorporating reinforcement learning and further validating the model against additional real-world scenarios.

\section{Discussion and Future Work}
The use of Transactional Analysis (TA) in Multi-Agent Systems (MAS) brings many new opportunities for creating agents with more psychological depth. This helps agents to interact more realistically by understanding individual needs and complex social signals. However, making these agents work in diverse and ever-changing environments is a challenge \cite{Gurcan}. The key issue is making sure the agents' behaviors remain true to established psychological principles. This requires careful validation of the consistency and accuracy of agents' actions \cite{Taverner, Argyle}. Keeping this psychological relevance is important to ensure that agents are reliable and can be used in practical applications \cite{Gao}.
\subsection{Architecture Improvements} Future improvements to Trans-ACT's architecture could include new features that will help agents learn, remember, and find better decisions in social situations. These improvements aim to make agents more psychologically advanced and better able to handle different social contexts.\\
\subsubsection{Reinforcement Learning} Reinforcement learning (RL) might be incorporated into Trans-ACT so that agents can learn from their experiences and adapt their behaviors over time. This could help them be more flexible in how they respond to social situations. Understanding how new experiences could influence the ego states \cite{Ceridono}, especially the Parent and Child states, which are shaped by early life experiences \cite{Stewart02}, is important. Adding RL could also allow the agents to change their life scripts over time, breaking free from old repetitive patterns and forming new ones based on what they learn from their environment. This would make agents more adaptable and closer to how human personalities evolve.\\
\subsubsection{Integration with Cognitive Architectures} Although Trans-ACT already has a memory system, we can improve it by connecting it with existing cognitive architectures such as ACT-R or Soar \cite{laird}. These cognitive systems have well-developed methods for reasoning, problem-solving, and learning. By combining Trans-ACT with these architectures, agents can simulate more complex social interactions and reason about situations more like humans do. However, this integration would need careful planning to ensure that the two systems work well together.\\
\subsubsection{More Advanced TA Connection} A useful next step would be adding more advanced Transactional Analysis concepts to Trans-ACT. Elements like \emph{drivers} (mini-script behaviors), \emph{racket feelings} (substitute emotions), \emph{strokes} (units of recognition or attention), and \emph{trading stamps} (accumulated unexpressed, negative feelings) can make the agents' behaviors more complex and closer to real human interactions. These elements represent deeper emotional patterns and unconscious motivations. For example, including stamp collecting could allow agents to simulate the build-up of frustration, which in turn could trigger disproportionately large emotional reactions.
\vspace{-10pt}
\subsection{Potential Real-World Applications} 
\vspace{-10pt}
The integration between Transactional Analysis and MAS offers numerous practical applications, particularly in areas where social cognition and psychological processes are crucial.\\
\subsubsection{Conflict Resolution and Mediation Tools} TA-based agents may assist in conflict resolution by understanding the psychological dynamics underlying disputes. TA-based agents could identify psychological patterns such as transactional games. Recognizing these patterns would allow agents to propose strategies to break the cycle and facilitate more constructive interactions. Agents could simulate ways to prevent or resolve conflicts in specific situations. This could lead to more effective mediation strategies in personal, organizational, and even diplomatic settings \cite{Torkaman, Ito, song}.\\
\subsubsection{Advancing Psychological Research} TA-based Multi-Agent Systems have the potential to advance psychological theory, particularly in understanding interpersonal dynamics and complex human behaviors. Social cognition theories on memory, emotion, and decision-making could be explored more deeply through these agent-based models. These systems can serve as valuable tools for developing and refining psychological knowledge \cite{Flache, Elston}. This could lead to a more comprehensive understanding of how early experiences shape behavior throughout life.\\
\subsubsection{Supporting Scientific Research in Social Systems} Another important application is using Trans-Act to enhance scientific research on social systems \cite{Ito, adasociety}. By modeling the interactions of cognitive agents, researchers could gain insight into how people form relationships, resolve conflicts, and navigate social hierarchies \cite{song, Flache}. These systems could simulate how groups with different psychological adaptations respond to social changes or policies. For example, simulations could predict how different populations might react to a new policy or social trend, helping to better understand the social impact of different decisions \cite{Argyle, Shapiro}. By incorporating psychological theories into agent-based models, we can simulate the cognitive and emotional responses that shape collective behavior, providing valuable data for social scientists and politics alike.

\section{Conclusion}
This study contributes to the growing field of cognitive agent modeling by demonstrating the potential of integrating Transactional Analysis with advanced cognitive mechanisms. The Trans-ACT framework structures agent behavior around distinct ego states (Parent, Adult and Child). By encoding long-term memories for each state and employing similarity-based retrieval, agents dynamically activate internal schemas that guide their responses in a way consistent with human cognition. This process not only reinforces behavioral consistency, but also prevents agents from "going in circles" by promoting progressive and adaptive interactions.

Experimental simulation, based on the \emph{Stupid} game, illustrates that agents with TA-inspired architectures can display rich psychological dynamics. Jordan's responses reveal a strong dependence on external validation and a tendency to avoid responsibility, while Alex consistently assumes control and attempts to help. The results confirm that the interplay between automatic emotional responses and deliberate analytical thinking plays a key role in shaping agent behavior. Moreover, the simulation highlights the importance of well-structured memories in triggering appropriate responses and ensuring that agents remain aligned with their internal life scripts.

We highlighted the broader applications of Trans-ACT across various domains. For example, in conflict resolution, the framework could model emotionally charged interactions to simulate strategies for deescalation \cite{deGraaf}. In education, agents might recreate realistic classroom dynamics to improve teacher training. In sociological research, Trans-ACT could help study collective behaviors or group decision-making processes \cite{Gurcan}. By enabling MAS to model complex psychological dynamics, this work offers new tools to address real-world challenges while promotes collaboration between cognitive science and LLMs.

\section{Acknowledgments}
The work reported in this paper was partly supported by the Polish National Science Centre under grant 2020/39/I/HS1/02861.

\bibliographystyle{apacite}

\setlength{\bibleftmargin}{.125in}
\setlength{\bibindent}{-\bibleftmargin}

\bibliography{article}

\end{document}